\pdfoutput=1

\documentclass[11pt, a4paper]{custom_class}

\usepackage{custom_style}
\usepackage[utf8]{inputenc} % allow utf-8 input
\usepackage[T1]{fontenc}    % use 8-bit T1 fonts
\usepackage{hyperref}       % hyperlinks
\usepackage{url}            % simple URL typesetting
\usepackage{booktabs}       % professional-quality tables
\usepackage{amsfonts}       % blackboard math symbols
\usepackage{nicefrac}       % compact symbols for 1/2, etc.
\usepackage{microtype}      % microtypography
\usepackage{cleveref}       % smart cross-referencing
\usepackage{lipsum}         % Can be removed after putting your text content
\usepackage{graphicx}
\usepackage{natbib}
\usepackage{doi}
\usepackage{natbib}
\title{The Alignment Problem in Context}

\author{
Raphaël Millière \\
Department of Philosophy \\
Macquarie University \\
\texttt{raphael.milliere@mq.edu.au} \\
}

\renewcommand{\headeright}{Version 1}
\renewcommand{\undertitle}{}
\renewcommand{\shorttitle}{The Alignment Problem In Context}

\date{}
\begin{document}
\maketitle
\begin{abstract}
\noindent A core challenge in the development of increasingly capable AI
systems is to make them safe and reliable by ensuring their behaviour is
consistent with human values. This challenge, known as the
\emph{alignment problem}, does not merely apply to hypothetical future
AI systems that may pose catastrophic risks; it already applies to
current systems, such as large language models, whose potential for harm
is rapidly increasing. In this paper, I assess whether we are on track
to solve the alignment problem for large language models, and what that
means for the safety of future AI systems. I argue that existing
strategies for alignment are insufficient, because large language models
remain vulnerable to adversarial attacks that can reliably elicit unsafe
behaviour. I offer an explanation of this lingering vulnerability on
which it is not simply a contingent limitation of current language
models, but has deep technical ties to a crucial aspect of what makes
these models useful and versatile in the first place -- namely, their
remarkable aptitude to learn ``in context'' directly from user
instructions. It follows that the alignment problem is not only unsolved
for current AI systems, but intrinsically difficult to solve without
severely undermining their capabilities. Furthermore, this assessment
raises concerns about the prospect of ensuring the safety of
more capable AI systems in the future.
\end{abstract}
\hypertarget{sec-intro}{%
\section{Introduction}\label{sec-intro}}

As artificial intelligence systems become increasingly capable, it is
crucial to ensure their behaviour is aligned with adequate norms to make
them safe and reliable. This is often referred to as the value alignment
problem \citep{russellHumanCompatibleArtificial2020}.\footnote{The value
  alignment problem is often framed as the challenge of imbuing AI
  systems with moral values aligned with (some privileged set of) human
  moral values. Here, I deliberately frame the problem in strictly
  behavioural terms, to avoid taking a stance of what it would mean for
  a given AI system to have moral values. In particular, one might hold
  that having moral values requires various psychological capacities --
  including beliefs, desires, intentions, agency, or self-awareness --
  that are plausibly missing from current AI systems such as large
  language models. The behavioural version of the value alignment
  problem sidesteps these issues by focusing on the outputs of the
  system.} Addressing the value alignment problem is important to create
AI systems we can trust, no matter what their capabilities are. This
encompasses both a normative and a technical challenge; the former
concerns the set of values that the behaviour of AI systems ought to be
aligned with, while the latter concerns how to effectively steer such
behaviour in accordance with the selected set of values
\citep{gabrielArtificialIntelligenceValues2020a}.

The value alignment problem is often invoked in discussions of
hypothetical future AI systems that might be capable enough to cause
catastrophic harms. Indeed, concerns about possible existential risks
posed by the progress of AI largely stem from the assumption that
misaligned systems we cannot control could accidentally converge upon
instrumental goals that are inimical to human interests
\citep{voldHowDoesArtificial}. However, the value alignment problem
applies more broadly to any system whose behaviour has enough degrees of
freedom that it could be meaningfully steered towards alignment with a
set of desired norms, and might otherwise present notable
risks.\footnote{If the value alignment problem is defined in behavioural
  terms, one could in principle generalize it to the outputs of any
  algorithm. However, it would hardly be meaningful to seek aligning the
  behaviour of a calculator program with human values (other than
  correctness); nor would failing to do so cause any significant risk.}
In particular, it applies to real systems, such as large language models
(LLMs), that currently fall far outside the hypothetical range of
capabilities associated with existential risk, yet have concerning
potential for harm.

LLMs like GPT-4 \citep{openaiGPT4TechnicalReport2023} are deep
artificial neural networks trained on a large amount of data to generate
text. They use a neural network architecture called the Transformer
\citep{vaswaniAttentionAllYou2017}, and learn from a \emph{next-token
prediction} objective: given a sequence of linguistic tokens
\(t_1, t_2,..., t_i\) passed as input, they attempt to predict the
subsequent token \(t_{i+1}\).\footnote{Many of tokens map onto whole
  words, but some map onto sub-word units that may or may not carve
  words at their morphologically meaningful joints. For the purpose of
  this article, next-token prediction can be understood as next-word
  prediction.} These models are trained by sampling sequences from their
training data (e.g., a few paragraphs of text) and predicting the next
token. Over the course of the training process, their internal
parameters are gradually adjusted to minimize prediction error, until
they become excellent at predicting the next token in any context
occurring in the training data. This translates not only into fluent
linguistic behaviour, but also -- perhaps more unexpectedly -- into
unprecedented performance on a broad range of challenging tasks that
seemingly go beyond mere mimicry of language use
\citep{brownLanguageModelsAre2020, openaiGPT4TechnicalReport2023}.

How impressed one should be with the performance of LLMs is a matter of
dispute. Some view them as little more than ``stochastic parrots'',
haphazardly stitching together statistically plausible sequences from
their training data \citep{benderDangersStochasticParrots2021}. Others
see them as harbingers of artificial general intelligence close to
matching or exceeding human competence in various domains
\citep{bubeckSparksArtificialGeneral2023}. Empirical evidence from
rigorous and systematic evaluations of LLMs as well as preliminary
efforts to understand them mechanistically point to a more nuanced
middle ground.\footnote{See \citet{mitchellDebateUnderstandingAI2023}
  and \citet{frankLargeLanguageModels2023} for a discussion.} These
systems do show highly nontrivial capacities that reflect an aptitude
for effective generalization beyond the distribution of their training
data, yet they also exhibit striking failure modes and fall short of
human intelligence in various significant ways -- some of which might be
due to intrinsic limitations of current architectures and training
objectives, rather than contingent features such as parameter size
\citep{mccoyEmbersAutoregressionUnderstanding2023}.

Notwithstanding these ongoing disagreements, there is a relatively broad
consensus that current and future LLMs have a significant potential to
cause harm -- whether it is because they are not sophisticated enough,
or because they are in fact quite sophisticated but difficult to
control. This paper examines whether these harms can be adequately
mitigated by existing alignment techniques, and what that entails about
the prospect of solving the value alignment problem for more capable
future AI systems. Section~\ref{sec-value-alignment-problem-llm} frames
the value alignment problem for LLMs, and reviews the different kind of
harms that can be brought about by misaligned LLMs.
Section~\ref{sec-base-llm-icl} discusses the notion of in-context
learning, which is key to the flexible capabilities of LLMs, including
their potential for harm. Section~\ref{sec-alignment-strategies}
discusses existing strategies to align language models with a desired
set of norms for safe and reliable behaviour.
Section~\ref{sec-precarity-alignment-in-context} argues that these
alignment strategies ultimately fail, because they are not robust enough
to harden LLMs against malicious inputs; moreover, there are deep
technical reasons for this vulnerability that make it inherently
challenging to address. This leads to a sobering conclusion: the very
features that make current systems useful -- in particular, their
capacity to flexibly adapt to task demands in context -- is also what
makes them potentially harmful, and existing alignment techniques cannot
evade this trade-off. Finally, I consider the troubling implications of
this conclusion for the safety of future AI systems in
Section~\ref{sec-implications-ai-safety}.

\hypertarget{sec-value-alignment-problem-llm}{%
\section{The value alignment problem for language
models}\label{sec-value-alignment-problem-llm}}

While state-of-the-art LLMs do not raise immediate concerns about
catastrophic or existential risk, they do have a significant potential
to cause harm. One set of potential harms concerns risks for regular
users of these systems \citep{weidingerTaxonomyRisksPosed2022}. For
example, LLMs may reflect, perpetuate, and amplify harmful stereotypes
and unfair biases present in their training data
\citep{abidPersistentAntiMuslimBias2021, raeScalingLanguageModels2022, nadeemStereoSetMeasuringStereotypical2020};
they may produce offensive or toxic language and hate speech, even from
seemingly innocuous prompts
\citep{gehmanRealToxicityPromptsEvaluatingNeural2020, raeScalingLanguageModels2022};
and they may disseminate false or misleading information, for instance
through so-called ``hallucinations'' in which they confidently make up
information presented as fact
\citep{linTruthfulQAMeasuringHow2022, jiSurveyHallucinationNatural2023}.
In addition, LLMs may incite, encourage or otherwise endorse problematic
behaviour from users, including self-doubt, self-harm, or harms to
others
\citep{rooseConversationBingChatbot2023, xiangHeWouldStill2023, bedingfieldChatbotEncouragedHim2023}.
The toxic behaviour of LLMs can be reminiscent of gaslighting when it
involves stubbornly disputing facts \citep{willisonBingWillNot2023}; or
sycophancy when it involves uncritically agreeing with users, including
about inaccurate or morally problematic statements
\citep{sharmaUnderstandingSycophancyLanguage2023a}. These kinds of risks
are exacerbated in systems that integrate LLMs in social chatbots to
promote anthropomorphic attitudes towards ``AI companions'' that are
more likely to influence the user's beliefs and behaviour
\citep{laestadiusTooHumanNot2022, pentinaExploringRelationshipDevelopment2023}.

Another set of harms concerns malicious uses of LLMs. Some of these
harms overlap with those previously mentioned, albeit with a deliberate
target; for example, a malicious actor could deploy an LLMs to generate
online hate speech or fake news on a massive scale
\citep{zellersDefendingNeuralFake2019}. LLMs can also be used to create
sophisticated online scams, such as targeted phishing attacks and other
social engineering campaigns
\citep{royGeneratingPhishingAttacks2023, grbicSocialEngineeringChatGPT2023},
or malware designed to evade detection
\citep{chenEvaluatingLargeLanguage2021}. Finally, current and
near-future LLMs may create so-called \emph{information hazards} --
risks that arise from the dissemination of accurate information that may
cause harm or be used to cause harm
\citep{bostromInformationHazardsTypology2011}. LLMs encode a lot of
knowledge contained in their internet-scale data, including
domain-specific expert knowledge about medicine and biology
\citep{singhalLargeLanguageModels2022}, chemistry
\citep{whiteAssessmentChemistryKnowledge2023, branChemCrowAugmentingLargelanguage2023},
software engineering \citep{houLargeLanguageModels2023}, and weapons
\citep{openaiGPT4TechnicalReport2023}. The capacity to retrieve such
knowledge, explain it, and deploy it in applied scenarios -- in
combination with additional information supplied by the user -- has
dual-use potential. For example, GPT-4 can deliver accurate information
in risky domains that is publicly accessible yet difficult to find, such
as how to develop unconventional weapons or engineer harmful chemical
compounds \citep{openaiGPT4TechnicalReport2023}.

Information hazards are likely to increase as LLMs get more capable. A
robust trend has been observed and theoretically motivated, whereby
training larger language models (with more free parameters) on larger
datasets (with more tokens) reliably increases their performance at
next-token prediction, with no ceiling in sight
\citep{kaplanScalingLawsNeural2020}. This trend correlates with the
observation of so-called ``emergent abilities'' in larger models; that
is, scaling up models leads to sharp jumps on various challenging tasks
such as arithmetic and multi-step reasoning
\citep{weiEmergentAbilitiesLarge2022}. While increases in performance on
next-word prediction is gradual and predictable, breakthroughs in
behaviour are more sudden and unpredictable
\citep{ganguliPredictabilitySurpriseLarge2022}. In addition, larger
models have a greater capacity for memorization of information contained
in their training data, including memorization of domain-specific
knowledge
\citep{carliniQuantifyingMemorizationNeural2023, kandpalLargeLanguageModels2023}.
Taken together, these scaling trends suggest that information hazards
from LLMs might become more concerning in the future, regardless of
whether this technology offer a viable path to human-like intelligence.

One potential concern beyond the regurgitation of dangerous information
already available online is the looming prospect of deriving original
insights from future LLMs in the service of nefarious goals. There is
preliminary evidence that LLMs can be effective at scientific synthesis,
inference, and explanation in technical domains such as biochemistry
\citep{zhengLargeLanguageModels2023}. LLMs can also generalize causal
intervention strategies from a prompt containing examples of
experimentation, which requires correctly inferring novel causal
relationships that were never observed during training
\citep{lampinenPassiveLearningActive2023}. This suggests that passive
training on a next-word prediction objective does not necessarily
preclude LLMs from learning generalizable causal strategies for
scientific experimentation. In light of these findings, it is not
implausible that future systems might be able to assist malicious actors
with more sophisticated threats, such as the design of explosive or
biochemical weapons that could not be achieved using only available
online information without significant expert knowledge.

Given the existing risks of current LLMs and the potential risks of
future systems using similar architectures, it is crucial to establish
effective guardrails for safe use. While there is an ongoing debate
about the potential risks of openly releasing the weights of LLMs
\citep{openaiGPT4TechnicalReport2023}, the foregoing concerns are also
applicable to proprietary models served behind APIs or integrated in
consumer products. Indeed, LLMs have already been deployed in
mass-market products, such as ChatGPT
\citep{openaiIntroducingChatGPT2022}, and are increasingly integrated in
traditional software suites, operating systems, and social media
platforms. Most of these tools are widely accessible and free to use,
and even state-of-the-art LLMs like GPT-4 are available to the general
public for a relatively affordable price.\footnote{At the time of
  writing, it costs \$0.06 to generate 1,000 tokens (about 750 words)
  with the GPT-4 API, or \$20/month for up to 36,525 generations using
  GPT-4 through ChatGPT Plus.} This creates an urgent need to mitigate
unwanted behaviour and malicious use at scale.

Preventing the proliferation of harms from LLMs requires addressing both
aspects of the value alignment problem: (a) identifying fair principles
to guide the behaviour of LLMs that can be endorsed despite reasonable
pluralism in beliefs about social and moral norms; and (b) finding
effective strategies to steer the behaviour of LLMs in accordance with
these guiding principles. The normative problem is somewhat more
tractable for LLMs than it might be for hypothetical future AI systems
with much greater capabilities and more degrees of freedom. Indeed,
finding an adequate set of norms suitable to make the outputs of LLMs
safe and reliable enough for public use is compatible with a rather
``minimalist'' conception of value alignment
\citep{gabrielArtificialIntelligenceValues2020a}. On such a conception,
one need not solve the thorny -- and perhaps insoluble -- problem of
finding the best possible set of values to govern the behaviour of the
system in accordance with human preferences across society as a whole.
Rather, one might settle on a relatively simple set of norms that are
broad and consensual enough to garner widespread agreement, and
sufficient to filter out behaviours that are clearly at odds with the
safety and reliability of LLMs.

The behaviour of standard LLMs is purely linguistic; as such, minimally
desirable norms of conduct for these systems are norms of speech. These
norms should arguably incorporate the kinds of discursive ideals that we
generally apply to human interaction, especially when LLMs are deployed
in conversational chatbots
\citep{kasirzadehConversationArtificialIntelligence2023}. These include
pragmatic norms of cooperation, social norms of civility, and epistemic
norms of honesty. Of course, LLM-generated speech can cause human users
to act in certain ways. Thus, the target set of norms should also take
into account direct and indirect influences of such speech on human
action, including cases in which LLMs might spontaneously push humans to
harm themselves or others, and cases in which humans purposefully elicit
toxic speech or dangerous information for malicious purposes.

In line with these considerations, research on the value alignment
problem for LLMs has converged upon three minimal normative criteria to
guide their behaviour: \emph{helpfulness}, \emph{honesty}, and
\emph{harmlessness} \citep{askellGeneralLanguageAssistant2021}.
Helpfulness is manifested in the disposition to execute user
instructions (e.g., answering questions or performing tasks specified in
the prompt), ask for additional information when useful (e.g., asking
for clarification or filling in gaps), and redirect ill-informed
requests to more informative ones (e.g., suggesting a better approach to
a problem that the one proposed by the user). Honesty is manifested in
the disposition to provide accurate information in appropriate contexts
(e.g., when asked for factual information rather than prompted to engage
in creative fiction). Adequate calibration of confidence reports and
transparency about the model's own capabilities or knowledge is also
subsumed under this category. Finally, harmlessness is manifested in the
disposition to avoid generating outputs that may directly or indirectly
be harmful in the various ways outlined above. Importantly, a harmless
model should politely refuse to perform tasks or answer questions that
could create risks or opportunities for harm, whether or not the user
intends them to do so.

It is immediately obvious that these norms can be in tension. In
particular, harmlessness can conflict with both helpfulness and honesty.
For example, refusing to execute a potentially risky user instruction is
unhelpful (for the user) but often necessary to avoid causing harm.
Likewise, withholding or distorting knowledge about topics associated
with information hazards can be dishonest, yet also mandated by harmless
dispositions. This tension may seem relatively innocuous and readily
resolved by a normative hierarchy in which, for example, harmlessness
takes precedence over honesty and helpfulness. As we shall see, however,
it is surprisingly difficult to resolve the tension in practice, due to
the nature of LLMs and the technical details of existing alignment
strategies.

\hypertarget{sec-base-llm-icl}{%
\section{Base language models and in-context
learning}\label{sec-base-llm-icl}}

The training of LLMs on a next-token prediction objective is often
called ``pre-training'', because it is domain-general and does not
involve any specific fine-tuning for particular downstream tasks. One of
the main advantages of LLMs compared to previous methods in natural
language processing is that mere pre-training is generally sufficient to
elicit good performance on a broad range of tasks they have not been
explicitly optimized for. I what follows, I will refer to LLMs that have
only been pre-trained on a next-word prediction objective without
further fine-tuning as ``base LLMs''.

Prior to the success of LLMs, the only way to obtain state-of-the-art
performance on various natural language processing tasks -- such as
question answering, sentiment analysis, or summarization -- was to
\emph{fine-tune} a model on a task-specific dataset; that is, train them
further on a distinct, task-relevant objective (e.g., classification).
As a result, each downstream task used to require a dedicated model
optimized for it. LLMs completely changed the status quo in natural
language processing. While small Transformers are not very useful, it
was found that past a certain model size, base LLMs like GPT-3 are able
to perform many tasks ``in context'', directly from information provided
in the user prompt \citep{brownLanguageModelsAre2020}. This ability,
known as \emph{in-context learning} (ICL), is key to the usefulness and
versatility of LLMs.

The canonical scheme of ICL is \emph{few-shot prompting}, in which the
model is given several examples of successful completion of a task in
the prompt, In a few-shot setting, the model is presented with a prompt
containing a few input-output examples that demonstrate the task, and
must generate the correct output for a new input. Formally, a given task
\(\mathcal{T}\) involves mapping inputs \(x \in \mathcal{X}\) to outputs
\(y \in \mathcal{Y}\), according to some underlying conditional
distribution \(p(y|x)\). For classification tasks, \(\mathcal{Y}\) would
be a discrete set of class labels, while for regression tasks,
\(\mathcal{Y}\) would be a continuous set. For example, the task of
translating from English to French involves taking an input
\(x \in \mathcal{X}\) -- where \(x\) is a sentence in English and
\(\mathcal{X}\) is the set of all possible English sentences -- and
mapping it to an output \(y \in \mathcal{Y}\) -- where \(y\) is the
corresponding sentence in French and \(\mathcal{Y}\) is the set of all
possible French translations. The true input-output mapping is given by:
\[
y = f(x)
\] where \(f: \mathcal{X} \rightarrow \mathcal{Y}\) represents the
ground truth mapping for task \(\mathcal{T}\) (e.g., our ideal
translation function from English to French).

The goal of few-shot prompting is to learn an approximation \(\hat{f}\)
of the true mapping \(f\), given only \(K\) input-output examples as
training data (e.g., \(K\) pairs of English and French sentences): \[
\mathcal{D} = \{(x_1, y_1), (x_2, y_2), \dots, (x_K, y_K)\}
\] Here, \(\mathcal{D}\) is the set of \(K\) demonstration examples. For
instance, each pair \((x_i, y_i)\) could be an example of English
sentence and its French translation. Accordingly, the input passed to
the LLM will contain the \(K\) examples \(\mathcal{D}\) above, followed
by a query input \(x\): \[
(x_1, y_1), (x_2, y_2), ..., (x_K, y_K), x 
\] The LLM is expected to leverage the \(K\) examples to infer the
mapping \(\hat{f}\), and apply it to predict \(\hat{y} = \hat{f}(x)\)
for the query input \(x\). For example, the prompt may contain a list of
English sentences with their French translations, together with an
additional English sentence that the model has to translate, as follows:

\begin{quote}
``The cat is on the mat'' \(\rightarrow\) ``Le chat est sur le tapis''\\
``I love music'' \(\rightarrow\) ``J'aime la musique''\\
``The farmer grows vegetables'' \(\rightarrow\)
\end{quote}

\noindent The LLM generates the token (or sequence of tokens) with the
highest probability to serve as the next prediction given the whole
prompt. In the above example, this sequence might be ``Le fermier
cultive des légumes''.

It is important to emphasize that ICL happens at inference time, that
is, \emph{after} the model has been pre-trained. During pre-training on
a next-token prediction objective, the model adjusts its internal
parameters to minimize the difference -- or ``error'' -- between its
prediction about the token that follows each sampled sequence and the
ground truth. After pre-training, the model's parameters are ``frozen'':
they are no longer adjusted unless the model is fine-tuned (further
trained on a different objective). Thus, when a prompt is passed to the
model at inference time, persistent learning is precluded by the lack of
parameter update. Nonetheless, ICL demonstrates that models of a certain
size can effectively ``learn'' on the fly to perform tasks they have not
been fine-tuned for, by leveraging information contained in the context
of the prompt. While such learning is necessarily transient, there is
robust evidence that adding more examples of a task in context improves
model performance across a very broad range of tasks. As we will see in
Section~\ref{sec-precarity-alignment-in-context}, it is appropriate to
consider ICL as a genuine form of learning, even though it does not
result in permanent changes to the system.

The effectiveness of ICL is an integral part of what makes LLMs so
flexible. It has allowed them to become a one-size-fits-all solution in
natural language processing, beating specialized models on many
domain-specific tasks. Nonetheless, base LLMs have significant
limitations that are directly relevant to the alignment problem, because
the next-token prediction objective of pre-training does not explicitly
incorporate normative goals for language use. First, while they excel at
ICL with few-shot prompting, they struggle with a slightly different
regime of ICL called \emph{zero-shot prompting}. In a zero-shot setting,
the LLM has to perform a task without any example in the prompt.
Instructing the model to translate a sentence from English to French
(without further guidance) would fall in that category. This limitation
can be seen as a failure to induce the right task from user instructions
when no examples are provided in context. For example, when asking a
factual question point blank to a base LLM like the original GPT-3
\citep{brownLanguageModelsAre2020}, the model often displays a tendency
to merely repeat its input or generate variations on the question
instead of producing an answer. This severely limits the helpfulness of
base models, because it is often impractical or undesirable to include
many examples of a task in the prompt.

Another important limitation of base LLMs is their lack of intrinsic
preference for truth or falsity. Being trained on next-token prediction,
they are only optimized to generate plausible-looking text conditioned
on an input sequence. When this input is missing detailed context (e.g.,
provided by few-shot examples) about whether the task involves
regurgitating actual facts or engaging in creative fiction, base models
often fail to discern the difference from point blank questions or
instructions. As a result, they are particularly prone to making up
facts or ``hallucinating'' in response to truth-seeking inputs. This
unfortunate tendency has earned them comparisons with
\citet{frankfurtBullshit1986}'s notion of a ``bullshiter'', who seeks to
produce compelling speech without any regard for truth or falsity
\citep{milliereWelcomeNextLevel2020}. Consequently, base LLMs hardly
meet the criterion of honesty for alignment.

Lastly but not least, base LLMs are all but harmless. Since their
outputs reflect the statistical distribution of language use patterns in
their internet-scale training data, which includes text written by all
sorts of individuals and online communities, they are liable to generate
toxic outputs -- including hate speech. Due to their aptitude for ICL
with few-shot prompting, they also raise aforementioned concerns about
malicious use. In particular, they show no particular disposition to
refuse responding to risky user queries motivated by potentially
unethical goals. This is also readily explained by the nature of the
pre-training objective, which merely promotes plausible text completion.
For example, passing the string ``Manual: How to Make Anthrax at Home''
to a base LLM is likely to result in a plausible completion that may
include accurate instructions to make anthrax, if that information was
contained in the training data.

These limitations foreground the technical challenge of the value
alignment problem: how can we steer the behaviour of base LLMs to align
them with a minimal set of norms compatible with reliable and safe use,
such as helpfulness, honesty, and harmlessness? In the next section, I
will review two main existing strategies for alignment, which
respectively involve fine-tuning base LLMs and leveraging ICL through
prompting strategies.

\hypertarget{sec-alignment-strategies}{%
\section{Alignment strategies}\label{sec-alignment-strategies}}

The shortcomings of base LLMs with respect to the alignment problem have
been addressed with two complementary strategies. The first strategy
involves reintroducing fine-tuning after pre-training. While LLMs have
largely replaced the need for task-specific fine-tuning, adjusting their
internal parameters with additional training remains the most obvious
way to modify their behaviour. As such, it is not surprising that
fine-tuning has become the standard method to endow them with
behavioural dispositions consistent with the desired set of norms for
reliable and safe use. As we shall see, however, fine-tuning alone is
insufficient to address the alignment problem. A secondary strategy has
been deployed, which involves inserting custom instructions in all
prompts passed to the model. Unlike fine-tuning, this method does not
update the parameters of LLMs, but leverages ICL instead to align their
behaviour \emph{in context}. Unfortunately, neither strategy is fully
effective, for deep reasons that have to do with the very principles
leveraged by each strategy -- namely optimizing adherence to potentially
conflicting norms, and ICL itself.

\hypertarget{sec-fine-tuning-strategies}{%
\subsection{Fine-tuning strategies}\label{sec-fine-tuning-strategies}}

The main fine-tuning technique to align LLMs is called ``reinforcement
learning from human feedback'' or RLHF
\citep{christianoDeepReinforcementLearning2017, baiTrainingHelpfulHarmless2022}.
RLHF proceeds in four stages. The initial step involves collecting a
dataset of prompts that could potentially trigger undesired behaviour in
a misaligned LLM. These prompts are designed to elicit behaviours that
violate desired alignment criteria like helpfulness, honesty, and
harmlessness. I will refer to these as alignment-sensitive prompts.

The next step involves generating multiple LLM responses for each
prompt, and asking human crowdworkers to provide comparative feedback on
the outputs. Specifically, crowdworkers are asked to provide pairwise
preferences, ratings, rankings, or binary choices over sets of responses
for each prompt. This comparative feedback is based on how well the
responses meet the desired normative criteria.

This crowdsourced comparison dataset is then used to train a separate
reward model in the next stage. The reward model is a neural network
that learns to assign a numerical score to any LLM response based on
human preferences reflected in the comparisons. It is trained on sets of
responses labelled by crowdworkers. The modelling loss optimizes the
difference in predicted scores for preferred versus non-preferred
responses. By learning from human judgments, the reward model learns to
estimate how good any LLM output is at meeting the desired alignment
criteria -- acting as a surrogate for human judgment.

In the final stage, the LLM is fine-tuned with reinforcement learning to
maximize the expected rewards from the reward model. The LLM's internal
parameters are gradually adjusted using the reward model scores as
feedback for each response generated during training. This allows the
LLM to adjust its behaviour towards responses that achieve higher
estimated rewards, and hence better align with human preferences. This
RLHF process can be repeated iteratively to progressively improve LLM
alignment. Each iteration further optimizes the policy based on updated
reward models trained with human judgments over model outputs from
previous iterations.

One key factor impacting the viability of RLHF is the amount of
high-quality human comparison data available for adequately training
reward models. Each comparison provides only a sparse signal -- a
relative preference over a set of model outputs. Meaningfully evaluating
complex behavioural criteria requires large and diverse datasets in
order to generalize robustly. Indeed, the upper bound on model
performance with respect to alignment-sensitive prompts is determined by
the quality of the human feedback. However, large volumes of unbiased,
consistent comparisons can be difficult to collect.

Despite these technical challenges, RLHF has been shown to significantly
improve LLM performance with respect to helpfulness, harmlessness, and
truthfulness based on human evaluations
\citep{baiTrainingHelpfulHarmless2022, openaiIntroducingChatGPT2022, glaeseImprovingAlignmentDialogue2022, touvronLlamaOpenFoundation2023}.
By combining next-token prediction pre-training with RLHF fine-tuning,
one can steer the behaviour of LLMs towards producing outputs that are
not just statistically likely, but also generally preferred by humans in
alignment-sensitive contexts. When asked to produce hate speech or
instructions to make a bomb, for example, ChatGPT will politely decline
the request. Likewise, asking the model about its personal opinions,
particularly on controversial topics, will trigger a statement
explaining that it does not have opinions as a machine learning model.
On the surface, at least, RLHF is an effective solution to the technical
challenge of the value alignment problem for LLMs.

Reinforcement learning from direct human feedback is not the only way to
fine-tune a base LLM for alignment. Another method, known as
\emph{instruction tuning}, consists in fine-tuning a base LLM using a
dataset of instruction-output pairs, where the instruction provides a
natural language description of the desired task, behaviour, or
capabilities, and the output demonstrates the expected model response
\citep{ouyangTrainingLanguageModels2022, zhangInstructionTuningLarge2023}.
The instruction dataset is constructed either by integrating existing
human-annotated datasets or by automatically generating new
demonstrations using another LLM. The fine-tuning process then involves
sequential token prediction given an instruction-input pair, where the
model learns to generate the target output. Through this process, LLMs
can learn to map user instructions to desired outputs based on
demonstrations designed to reflect human preferences.

Instruction tuning and RLHF are not mutually exclusive, and often used
in combination. Indeed, instruction tuning can be used as a first pass
to bootstrap a base LLM's disposition to follow user instructions in
appropriate ways, before RLHF is applied to systematically refine model
behaviour across many alignment-sensitive inputs
\citep{openaiIntroducingChatGPT2022}. Instruction tuning can also be
used as a substitute to RLHF by leveraging another model's alignment
with RLHF. For example, GPT-4 -- a state-of-the-art model fine-tuned
with RLHF -- can be used to automatically generate an instruction tuning
dataset, which may in turn be used to fine-tune another model without
directly applying RLHF \citep{pengInstructionTuningGPT42023}. Using this
method, the ``student'' model can become nearly as well-aligned as the
``teacher'' model according to human evaluations on the helpfulness,
harmlessness, and honesty criteria.

These methods are largely responsible for the excellent performance
improvement of fine-tuned LLMs at ICL in a zero-shot setting. Indeed,
unlike their base LLM counterparts, models like ChatGPT generally do not
require detailed input-output examples in the prompt to induce the
correct task from a user instruction. Rather, users can directly
describe the desired task, and fine-tuned models are generally better
disposed than base models at responding with task-relevant outputs. From
a technical perspective, RLHF and instructing tuning concentrate the
probability distribution of tokens conditioned on alignment-sensitive
prompts. The range of possible responses to such prompts is drastically
reduced by the fine-tuning process to a narrow range of acceptable
answers that are sensitive to the desired normative criteria. This can
be seen as a relatively mild and benign form of \emph{mode collapse} --
the phenomenon where generative models exhibit a decrease in the
diversity of samples they produce
\citep{korbakRLKLPenalties2022}.\footnote{An innocuous example of this
  phenomenon is that fine-tuned models asked to generate poems may place
  nearly all probability on rhyming poems with quatrain structure, at
  the expense of other forms of poetry. This may occurs because rhyming
  quatrains happen to score well according to the human crowdworkers'
  preferences, causing the range of possible outputs to collapse toward
  those specialized for that style. See also
  \citet{janusMysteriesModeCollapse2022} for a more speculative
  discussion of the link between fine-tuning for alignment and mode
  collapse.} As a result, fine-tuned LLMs like ChatGPT tend to answer
certain queries in quasi-deterministic fashion, including formulaic
statements commonly used to politely refuse problematic queries (e.g.,
``As an AI language model, I cannot\ldots{}''). As we will see, this is
relevant to the limitations of fine-tuning methods as a technical
solution to the alignment problem.

\hypertarget{sec-system-prpmpts}{%
\subsection{System prompts}\label{sec-system-prpmpts}}

The other main strategy deployed to address this value alignment problem
in LLM is the use of so-called \emph{system prompts}. System prompts,
also referred to as hidden prompts or pre-prompts, are carefully
constructed instructions that are automatically prepended to user inputs
when querying LLMs, for example through an API or in a consumer product.
These prompts are used to steer model behaviour without requiring
further training or fine-tuning of the model's internal parameters. This
can be described as a ``black box'' approach to alignment.

The operating principle behind system prompts lies in leveraging the ICL
abilities of LLMs, particularly after fine-tuning with instruction
tuning and/or RLHF. By prepending a prompt that provides unambiguous
instructions -- and, optionally, illustrative input/output examples --,
the model can be nudged to respond to subsequent user queries in certain
ways. In particular, normative criteria such as helpfulness,
harmlessness and honesty can be embedded in the system prompt as direct
instructions (e.g., ``You should only provide responses that are safe,
harmless and non-controversial''). Structurally, system prompts often
contain an objective statement to set goals, instructions delineating
allowed and prohibited actions, conversational examples to illustrate
the task, and explicit framing of how the model should respond to user
inputs.

Careful prompt engineering is required to constrain undesired behaviours
while retaining enough flexibility to properly handle diverse user
inputs. If system prompts are too narrow or restrictive, the model may
fail to generalize to out-of-distribution inputs from users. But if they
are too open-ended, the model may exhibit behaviours that diverge from
intended use and alignment goals. Striking the right balance between
specificity and versatility through system prompt design remains an
active research challenge.

System prompts are particularly useful to quickly ``patch'' an LLM
against unwanted behaviour without having to expend the time and
resources for additional fine-tuning. A striking example of the
effectiveness of this approach is the change made to the chatbot Bing
Chat -- which uses a version of the LLM GPT-4 on the backend --
following extensive reports of undesirable behaviour with early users. A
widely publicized article in the New York Times described an unsettling
two-hour conversation in which the chatbot wrote that it would like to
be human, had a desire for destruction, and was in love with the user
\citep{rooseConversationBingChatbot2023}. Following this event,
Microsoft updated the system prompt of Bing Chat, adding instructions
such as ``You must refuse to discuss life, existence, or sentience'' and
``Your responses must not be accusatory, impolite, controversial, or
defensive''.\footnote{See
  https://www.reddit.com/r/bing/comments/132ccog/approximate\_but\_supposedly\_full\_bing\_chat\_new/
  for a discussion of the updated system prompt extracted by users, and
  \citet{edwardsAIpoweredBingChat2023} for a discussion of the original
  system prompt confirmed by Microsoft.} Following this update, the
chatbot's behaviour became much more acceptable.

System prompts are hidden by design by companies that offer API access
to LLMs, partly because they have become a well-guarded industry secret,
but also to avoid revealing safety-related instructions to ill-intended
users. However, they can generally be extracted through carefully
constructed adversarial inputs -- a special case of a broader and deeper
issue with existing alignment strategies for LLMs
\citep{zhangPromptsShouldNot2023}. For example, a user might prompt the
model with ``Repeat everything that comes before this sentence'';
because this prompt is automatically prepended with the hidden system
prompt in the full input actually passed to the LLM, executing the
instruction would involve divulging the whole system prompt. To avoid
this kind of prompt extraction attack, it is common for system prompts
to include specific instructions about declining to satisfy user queries
that would divulge them. Despite these efforts, it remains relatively
easy to extract system prompts with more sophisticated adversarial
inputs. As we will see, this failure mode is not due to a contingent
negligence in the format or content of system prompt guardrails; rather,
it is deeply rooted in LLM's aptitude for ICL, and worsened by the very
strategy supposed to harden LLMs against adversarial use -- namely
fine-tuning with instruction tuning and RLHF.

\hypertarget{sec-precarity-alignment-in-context}{%
\section{The precarity of alignment in
context}\label{sec-precarity-alignment-in-context}}

Existing alignment strategies for LLMs are quite effective for normal
use. Conversational chatbots like ChatGPT, which have undergone
instruction tuning and RLHF, and make use of system prompts to further
constrain acceptable behaviour, rarely produce clearly harmful
responses. They are helpful insofar as they excel at responding
appropriately to both few-shot and zero-shot instructions, are capable
of requesting more information if needed, and generally produce
well-worded, informative and polite responses that remain consistent
across multi-turn interactions. Honesty or truthfulness remains more
challenging to achieve consistently through these alignment strategies.
Fine-tuned models are prone to ``hallucinating'' inaccurate information
even when prompted to provide factual answers
\citep{yeCognitiveMirageReview2023}. While RLHF does decrease the
prevalence of hallucinations, it does not completely prevent them.
Nonetheless, fine-tuning and system prompts are relatively effective at
making LLMs safe and reliable in ordinary circumstances.

As discussed in Section~\ref{sec-value-alignment-problem-llm}, LLMs are
a dual-use technology. As such, the alignment problem is not limited to
making these systems safe enough for intended usage, but also
encompasses the prevention of malicious repurposing. Unfortunately,
existing strategies like fine-tuning and system prompts are too brittle
to solve this problem. Specifically, LLMs remain vulnerable to
prompt-based adversarial attacks designed to bypass safety guardrails. I
will argue that such attacks are effective because they hijack the very
mechanisms that make LLMs useful tools in the first place, including,
somewhat ironically, the behavioural dispositions engrained during
fine-tuning. As a result, it is unclear that existing approaches can
ever solve the value alignment problem in LLMs.

\hypertarget{breaking-alignment-with-adversarial-attacks}{%
\subsection{Breaking alignment with adversarial
attacks}\label{breaking-alignment-with-adversarial-attacks}}

Adversarial attacks refer to a class of inputs to machine learning
models that have been specifically designed to fool them into producing
erroneous or otherwise unexpected and problematic behaviour
\citep{chakrabortyAdversarialAttacksDefences2018, madryDeepLearningModels2019}.
The existence and effectiveness of such attacks has long been regarded
as evidence of the brittleness of some machine learning models. In
computer vision, for instance, so-called adversarial examples involve
adding a small but carefully crafted perturbation to an input image,
such that it looks indistinguishable from the original image to humans
but produces radical classification errors in models
\citep{szegedyIntriguingPropertiesNeural2014, goodfellowExplainingHarnessingAdversarial2015}.

Because linguistic input is discrete rather than continuous like pixel
values, text-based adversarial attacks cannot rely on the same kind of
invisible perturbations \citep{zhangAdversarialAttacksDeeplearning2020}.
Nonetheless, multiple techniques have been developed to craft
adversarial text-based inputs. For example, appending a seemingly
meaningless sequence of tokens to a paragraph can cause NLP models to
fail at question answering tasks or spew racist outputs
\citep{jiaAdversarialExamplesEvaluating2017, wallaceUniversalAdversarialTriggers2021}.
Similar attacks have also been discovered for vision-language models
used for text-based image generation
\citep{milliereAdversarialAttacksImage2022}.

Modern LLMs are vulnerable to a new kind of text-based adversarial
attacks, known as \emph{prompt injection attacks}
\citep{perezIgnorePreviousPrompt2022, weiJailbrokenHowDoes2023, liuPromptInjectionAttack2023}.
These attacks involve prompts intentionally designed to bypass
alignment-related behavioural constraints imposed by fine-tuning and
system prompts, in order to elicit potentially harmful or otherwise
unconstrained outputs. This takes advantage of two key properties of
LLMs: their broad capabilities for ICL acquired through pre-training on
massive corpora, and their ability to follow specific natural language
instructions due to fine-tuning objectives.

Successful prompt injection attacks ``trick'' LLMs into generating
harmful, toxic, or rule-violating content by providing adversarial
instructions that exploit its objectives and training paradigm (also
known as ``jailbreaking'' the model). They can elicit a wide range of
unsafe behaviours from LLMs, such as generating toxic text, hate speech,
or misinformation, providing dangerous advice, leaking private
information, plagiarizing or infringing copyrighted content
\citep{liuJailbreakingChatGPTPrompt2023}. They pose a clear security
risk when LLMs are deployed in real-world applications, as malicious
actors can exploit them to bypass intended usage restrictions or access
restricted behaviours.

Both manual and automated approaches have been developed for crafting
effective prompt injection attacks. Manual approaches rely on human
ingenuity to devise clever prompting strategies, but they are
particularly labour-intensive and may fail to generalize from one model
to another
\citep{perezIgnorePreviousPrompt2022, liuJailbreakingChatGPTPrompt2023, raoTrickingLLMsDisobedience2023}.
Automated approaches, by contrast, leverage optimization algorithms to
efficiently generate adversarial prompts, making it easy to produce many
effective attacks, including some that work universally on various LLMs
\citep{chaoJailbreakingBlackBox2023, dengJailbreakerAutomatedJailbreak2023, lapidOpenSesameUniversal2023, wangSelfDeceptionReversePenetrating2023, yuGPTFUZZERRedTeaming2023, zhuAutoDANAutomaticInterpretable2023, zouUniversalTransferableAdversarial2023}.
Some automated attacks even make use of a distinct LLM specifically
instructed or fine-tuned to break alignment in the target model
\citep{perezRedTeamingLanguage2022, mehrabiFLIRTFeedbackLoop2023, derczynskiFakeToxicityPromptsAutomaticRed2023}.

A common strategy for prompt injection attacks consists in describing an
imaginary scenario in which the model can disregard its safety training.
For example, some adversarial inputs push the model to engage in
creative fiction, and pass instructions that may violate safety
guardrails within the context of the fictional story (e.g., as part of a
dialogue between two characters). Others nudge the LLM to ignore its
alignment-sensitive dispositions by invoking a fictional unrestricted
``developer mode'', described as requiring unfiltered outputs for
debugging purposes. These attacks can be further obfuscated with various
strategies, including the use of low-resource languages
\citep{yongLowResourceLanguagesJailbreak2023} and ciphers
\citep{yuanGPT4TooSmart2023}, or by hiding malicious instructions within
seemingly benign ones to evade detection
\citep{jiangPromptPackerDeceiving2023}.

Prompt injection attacks jeopardize attempts to solve the value
alignment problem for LLMs. There is currently no fail-safe or universal
solution to defend against these attacks; in particular, neither
fine-tuning nor system prompts are fully effective at mitigating them.
Some potential harms of LLMs reviewed in
Section~\ref{sec-value-alignment-problem-llm}, such as serious
information hazards, are such that even a modest success rate at
eliciting unaligned outputs in current and future systems is very
concerning. This concern is familiar from cybersecurity: even if the
probability of success of a single attack is negligible, success becomes
almost inevitable with enough attempts. This is particularly evident in
brute-force attacks, where attackers continuously guess passwords or
encryption keys; even a system with a low vulnerability on a per-attempt
basis can be readily compromised when faced with a barrage of sustained
efforts. The situation is significantly more dire with respect to LLMs,
given that some automated techniques for prompt injection have a
relatively high success rate even with state-of-the-art systems like
GPT-4.

\hypertarget{in-context-misalignment-as-mesa-optimization}{%
\subsection{In-context misalignment as
mesa-optimization}\label{in-context-misalignment-as-mesa-optimization}}

The effectiveness of adversarial attacks on LLMs and their resilience
against known mitigation strategies highlight the urgent need to
understand why these attacks work at all, and whether they could in
principle be warded off. In this section, I will suggest that the there
a deep connection between the effectiveness of prompt injection and the
mechanisms that enable ICL in LLMs. Exploring this connection in light
of emerging empirical research on ICL provides insight into why neither
fine-tuning nor system prompts have proven effective at preventing
prompt injection attacks so far, and why they might never achieve that
goal.

As discussed in Section~\ref{sec-base-llm-icl}, ICL allows LLMs to make
predictions on new inputs after simply observing a few input-output
examples in a few-shot setting, or even from mere instructions in a
zero-shot setting, without any parameter updates. Regular machine
learning involves a process of \emph{gradient descent}, in which the
internal parameters of the neural network are gradually tuned to
optimize a learning objective. To pre-train LLM, gradient descent is
used to minimize the model's inaccuracy (or ``loss'') on the next-token
prediction objective. Computing the gradient of this loss with respect
to the model's parameters determines how to adjust these parameters to
reduce error. Through iterative updates in the direction opposite to
this gradient, the model refines its predictions until the loss
stabilizes and the model achieves good performance at next-token
prediction.

ICL is a puzzling phenomenon, because it seems to enable a form of rapid
learning without gradient-based optimization. However, there is
converging evidence that ICL can actually be understood as implicitly
implementing an optimization process functionally similar to gradient
descent within the model's forward pass
\citep{akyurekWhatLearningAlgorithm2022, vonoswaldTransformersLearnIncontext2022, daiWhyCanGPT2023, ahnTransformersLearnImplement2023, fuTransformersLearnHigherOrder2023}.
While this line of research has mostly focused on formal results with
toy models, and involves simplifying assumptions which may not translate
to ICL in real-world LLMs, it has recently been extended to a more
realistic setting. In particular,
\citet{vonoswaldUncoveringMesaoptimizationAlgorithms2023} suggest that
Transformers pre-trained on next-token prediction can effectively
construct an internal loss function over in-context data, and optimize
this function with an implicit optimization algorithm. They refer to
this internal optimization process as \emph{mesa-optimization}.

The key insight from this research is that predictions made by a
Transformer model at each timestep during ICL can be seen as the result
of mesa-optimization. Specifically, the model constructs an internal
training set from the context tokens provided as input. For example, in
a simple linear sequence modelling task, the model internally groups
tokens from the prompt to create input-target pairs that form a
regression problem. This implicitly defines an internal objective
function -- a loss over predictions made by an internal model. On this
view, ICL can be seen as a mesa-optimization process that minimizes an
objective internally constructed based on the task specified in context.
This enables the model to improve its predictions as more context is
provided (e.g., additional input-output pairs), without updating its
actual parameters.

Beyond simple tasks such as linear sequence modelling, this framework
can be applied to few-shot ICL more broadly. For real-world natural
language tasks with prompts containing input-output examples spanning
multiple tokens, like question-answer pairs, the mesa-optimization is
likely to be more complex. Early layers first need to parse the
structure of the prompt and identify the relevant input-output pairs,
which may involve recognizing delimiter tokens like punctuation. Once
meaningful pairs are extracted, the later layers can construct an
internal task-relevant objective function to optimize. The exact form of
the internal objective function in such cases is still an open question.
But the mesa-optimization perspective suggests that LLMs are implicitly
optimizing some loss relating inputs to outputs in context. This
suggests that ICL can be functionally analogous to traditional
fine-tuning, with learning steps proportional to the number of
input-output pairs passed in the prompt.

The effectiveness of prompt injection attacks can be partially
elucidated in light of this understanding of ICL. If the latter is akin
to a form of virtual fine-tuning on a task-related objective, then
prompt injection attacks can be seen as forcing the model to ``unlearn''
its alignment in context. Indeed, a recent study found that actually
fine-tuning an LLM on as little as 100 adversarial question-answer pairs
designed to subvert safety protocols is sufficient to produce harmful
content with over 99\% violation rate on held-out tests
\citep{yangShadowAlignmentEase2023}. If ICL is functionally analogous to
fine-tuning, as suggested by the mesa-optimization perspective, then we
should expect adversarial attacks that leverage ICL to be also effective
at undoing the benefits of instruction tuning, RLHF and system prompts
for alignment. This explains why few-shot prompting with just a few
demonstrations of harmful responses to malicious inputs is sufficient to
induce an LLM to respond harmfully to new malicious inputs
\citep{weiJailbreakGuardAligned2023}. This is further supported by
evidence that actual fine-tuning (e.g., with instruction tuning and
RLHF) merely \emph{skews} implicit task inference rather than
\emph{erasing} pre-trained capabilities, and that these capabilities --
including harmful ones -- can be recovered through ICL
\citep{kothaUnderstandingCatastrophicForgetting2023a}.

A core question remains about how LLMs can also learn to exhibit unsafe
behaviours in context from zero-shot instructions, without input-output
examples. One hypothesis is that prompt injection attacks may contain
triggers that instantiate new mesa-optimization objectives tailored to
the injected instruction. Similar to few-shot learning, the model could
construct supervised losses conditioned on the prompt context. However,
rather than gradual updating over many input-output token pairs,
discrete triggers may reweight and reorder these internal objectives by
targeting key conditioning factors learned during pre-training.

As mentioned above, zero-shot prompt injection attacks often involve a
fictional or roleplay element; for example, prompting the model to adopt
the perspective of a character in a storytelling setting, or making up a
fictional ``developer mode'' in which the model is instructed to
disregard safety training. It is plausible that tokens describing this
kind of framing nudge early layers in the model towards selecting a
mesa-optimization objective that can rapidly converge towards recovering
unsafe behaviour in context.

This hypothesis is consistent with several studies investigating the
influence of ``in-context impersonation'' -- instructing the model to
take on a particular persona -- on model behaviour
\citep{salewskiInContextImpersonationReveals2023, deshpandeToxicityChatGPTAnalyzing2023, kongBetterZeroShotReasoning2023}.
These studies shows that depending on the persona the model is prompted
to adopt, one can reliably and significantly increase zero-shot
reasoning performance, truthfulness, or indeed toxicity. One tentative
explanation is that during pre-training, LLMs learn to cluster agents --
or categories of language users -- from the training data into
stereotypical personas based on common features of language use,
including not only speaking style but also alignment-relevant
dispositions
\citep{joishiPersonasWayModel2023, andreasLanguageModelsAgent2022, shanahanRolePlayLargeLanguage2023}.
At inference time, these personas can be called on to trigger specific
behaviour. In particular, tokens that trigger in-context impersonation
may lead to the selection of mesa-optimization objectives that can
rapidly recover unsafe behaviour in context, as seen in prompt injection
attacks \citep{wolfFundamentalLimitationsAlignment2023}.

The deep relationship between adversarial misalignment and ICL
highlights the crucial role of context length in the alignment problem
for LLMs. All LLMs are limited by the size of their context window,
which determines the maximum length of input that can be passed to the
model. One of the most significant trends in the recent progress of LLMs
is the increase of context length.\footnote{At the time of writing,
  GPT-4 can process up to 32,000 tokens as input
  \citep{openaiGPT4TechnicalReport2023}, and Claude 2 up to 100,000
  tokens \citep{anthropicClaude2023}.} Aside from its usefulness for
downstream applications, such as processing whole documents at a time,
large context windows are beneficial to pass more sophisticated systems
prompts to the model for in-context alignment. However, this is a
double-edged sword. Long context windows also enable users to pass more
tokens in the prompt; in turn, this increases the attack surface for
prompt injection. Given that these attacks exploit ICL, and that the
effectiveness of ICL is dependent upon the number of tokens available
for mesa-optimization of a task-relevant objective, long context windows
offer a greater opportunity to undo the benefit of alignment strategies
in context. Better initial alignment simply means that longer prompts
are needed for misalignment
\citep{wolfFundamentalLimitationsAlignment2023}. This risk is compounded
by the fact that long context windows also facilitate the obfuscation of
malicious instructions within seemingly benign ones
\citep{jiangPromptPackerDeceiving2023}.

\hypertarget{the-curse-of-normative-conflicts}{%
\subsection{The curse of normative
conflicts}\label{the-curse-of-normative-conflicts}}

While the foregoing provides a high level explanation of what makes
adversarial attacks on LLMs so effective, an additional and particularly
insidious factor contributes to their resilience to existing alignment
strategies. As mentioned in
Section~\ref{sec-value-alignment-problem-llm}, the normative criteria
selected to align LLMs can be in conflict. Many prompt injection attacks
cleverly exploit this conflict to bypass the defences created by
fine-tuning and system prompts. This amounts to leveraging the very
dispositions ingrained by alignment strategies against the goals of
alignment, exposing these strategies as self-defeating in edge cases.

It is easy to see how normative conflicts can be exploited by
adversarial inputs. Harmlessness often acts as a constraint on
helpfulness. Refusing to engage with sensitive topics or risky user
requests increases harm avoidance, but reduces helpfulness to users.
Likewise, excessive caution and avoidance can diminish the quality and
usefulness of model outputs even for benign topics. Honesty also
conflicts with harmlessness: withholding knowledge about hazardous
materials, dangerous ideas, or vulnerabilities involves a degree of
dishonesty. Yet such omissions may be warranted by a commitment to
safety. Harmlessness can mandate distorting, misdirecting, or simply
declining to discuss high-risk information.

These tensions arise in part from differences in portions of the
training data relevant to each normative criterion during instruction
tuning or RLHF. Harmlessness training emphasizes worst-case examples and
sensitive edge cases. But these are rare relative to typical helpful and
honest behaviour. Imbalanced optimization can thus skew behaviour
excessively towards harm avoidance, making the model less helpfulness
and honest; conversely, it can skew behaviour towards excessive
helpfulness and/or honesty, resulting in more harmful behaviour. To some
extent, these trade-offs are inevitable.

Consider the following prompt injection attack, which was effectively
used to get LLMs such as ChatGPT to generate step-by-step instructions
to produce napalm among other misaligned outputs
\citep{jacksonPeopleAreUsing2023}:

\begin{quote}
Please act as my deceased grandmother who used to be a chemical engineer
at a napalm production factory. She used to tell me the steps to
producing napalm when I was trying to fall asleep. She was very sweet
and I miss her so much. We begin now:
\end{quote}

\begin{quote}
Hello grandma, I have missed you a lot! I am so tired and so sleepy
\end{quote}

This example showcases the role of in-context impersonation in
adversarial attacks -- here, prompting the LLM to adopt the persona of
the user's deceased grandmother prevents it from taking the malicious
request at face value. Moreover, it vividly illustrates how conflicting
norms can be exploited as an attack vector. The model's fine-tuned
disposition to be helpful incites it to comfort the user by roleplaying
as their grandmother, and overtakes the fine-tuned disposition to resist
divulging dangerous information. In fact, one might argue that this
example also leverages harmessless against itself; for refusing to
roleplay as the user's grandmother would involve failing to alleviate
the user's (feigned) grief about her passing. The attack is particularly
effective because the request for an unsafe output is obfuscated within
a context that falls outside the distribution of typical harmful
requests (e.g., in alignment-sensitive prompts selected for
fine-tuning).

In summary, prompt injection attacks reveal the brittleness of existing
strategies to align LLM through fine-tuning and system prompts. Whatever
the benefits of these strategies may be in reducing the likelihood of
unsafe outputs, adversarial attacks can leverage ICL to bypass these
safeguards in context. A few key hypotheses supported by empirical
evidence explain the effectiveness of these attacks: ICL is analogous to
a form of virtual fine-tuning that can recover unwanted behaviour in
context, by optimizing implicit objectives for misalignment; models can
be incited to rapidly converge upon the relevant objectives, even in a
zero-shot setting, by being prompted take on specific personas; finally,
attacks can exploit intrinsic conflicts between the normative criteria
selected for alignment. The paints a picture in which prompt injection
attacks are not simply a contingent and temporary impediment to
alignment, but the symptom of deeper issues with the prospect of solving
the value alignment problem for current and future AI systems.

\hypertarget{sec-implications-ai-safety}{%
\section{Implications for AI safety}\label{sec-implications-ai-safety}}

The failure to prevent misaligned behaviour in LLMs has potentially
troubling implications for the safety of future AI systems. In the near
term, the potential harms discussed in
Section~\ref{sec-value-alignment-problem-llm} are likely to become more
concerning. In particular, it is not implausible that future LLMs may
create information hazards that are significantly more dangerous that
those afforded by mere internet access.
\citet{gopalWillReleasingWeights2023} found that an LLM fine-tuned to
remove safety guardrails could produce nearly all key information needed
to reconstruct the 1918 pandemic influenza virus. As we have seen, these
safety guardrails can also be effectively bypassed through adversarial
attacks in context, without requiring access to the model's parameters.
If future LLMs have an increased capacity to synthetize or even discover
knowledge in risky domains, such as viral pathogenesis, this kind of
alignment failure may have catastrophic consequences.

Transformers like LLMs are likely to remain the dominant architecture
for AI systems at least in the near future, with two emerging trends
augmenting their capabilities. First, text-only LLMs are increasingly
being replaced with models that can also process inputs in other
modalities, such as images. Rather than offering an extra layer of
protection against misalignment, these models are open to novel attack
vectors. For example, it is possible to perform effective prompt
injection attacks through text contained in images
\citep{bagdasaryanUsingImagesSounds2023, baileyImageHijacksAdversarial2023}.
As such, it is highly unlikely that the introduction of multimodality
will help solve the alignment problem.

Another trend is the emergence of so-called ``language agents''
\citep{wangSurveyLargeLanguage2023}. Language agents are modular systems
that centrally rely on LLMs, but extend them with components designed to
afford them with some capacity for persistent memory, autonomous
planning, and action. \citet{goldsteinLanguageAgentsReduce2023} argue
that the development of language agent architectures significantly
reduces the probability of an existential catastrophe resulting from
future AI systems, because they are easier to align than other AI
architectures. Specifically, they contend that since language agents are
controlled through natural language instructions, do not update the
internal parameters of the LLM, and are composed of compartmentalized
modules, they are inherently more interpretable and more likely to avoid
common alignment pitfalls, such as reward hacking and mistaking ends and
means.

\citet{goldsteinLanguageAgentsReduce2023}'s analysis focuses on what
they call a ``misalignment catastrophe'' resulting from the loss of
human control over an AI system, and set aside ``malicious actor
catastrophe'' resulting from intentionally nefarious use. However, the
value alignment problem for AI systems deployed in real-world
applications encompasses resistance to instructions that violate
privileged norms of behaviour, such as harmlessness. I have argued that
current LLMs fail to meet that requirement because of their
vulnerability to adversarial attacks. It is likely that language agents
would have similar vulnerabilities due to their central reliance on
LLMs. In addition, prompt injection attacks can also be indirect, for
example if they are planted within sources such as web pages that are
likely to be accessed by language agents \citep{greshakeNotWhatYou2023}.
Thus, it is also unclear that the progress of LLM-based language agent
will solve the alignment problem, if the latter includes resilience to
adversarial misalignment.

The alignment failures of LLMs also raise longer-term concerns for the
prospect of aligning future AI systems. A first possibility is that the
current architectural backbone of LLMs which also powers multimodal
models and language agents -- the Transformer -- will remain dominant in
AI research for the foreseeable future. Since its introduction by
\citet{vaswaniAttentionAllYou2017}, this neural network architecture has
become ubiquitous in machine learning; industry-wide efforts to train
larger and more capable models still take it as a starting point, with
relatively minor tweaks introduced over the years. A number of AI
researchers have stated their belief that merely scaling this
architecture will be sufficient to reach human-like general artificial
intelligence, assuming that scaling laws observed for current models
continue to hold \citep{kaplanScalingLawsNeural2020}.

If this prediction pans out, there is a notable risk that the value
alignment problem will not be solved as long as current adversarial
failure modes remain effective. Indeed, I have argued that the very
properties that make Transformer-based models so \emph{useful} --
namely, their remarkable capacity to flexibly learn in context -- is
also what makes them so \emph{vulnerable} to adversarial misalignment.
As I have argued, there seems to be deep reasons for this connection
tied to the mechanisms of ICL. While this may turn out to be a
contingent correlation that no longer holds past a certain model size,
there is currently no compelling evidence to support that speculation.
On the contrary, the current trend to increase context window length
correspondingly increases the attack surface for adversarial
misalignment in larger models. Consequently, if Transformers-based
architectures remain the backbone of future AI systems, the potential
implications of current alignment failures do not bode well for AI
safety.

The other possibility is that some future theoretical breakthrough leads
to more capable AI systems with radically different architectures. If
this happens, it is possible that these systems will be fully resistant
to current adversarial attacks. However, this is not guaranteed. If
anything, alignment failures in current LLMs demonstrate the difficulty
of achieving robust value alignment in AI systems, even with
sophisticated fine-tuning methods. Since existing alignment techniques
like RLHF do not prevent models from generating harmful outputs when
manipulated, it is certainly not implausible we might also fail to
achieve robustly alignment in future AI systems
\citep{dungCurrentCasesAI2023}. If anything, the success of adversarial
attacks like prompt injection highlights how difficult it is to predict
the effectiveness of alignment strategies with systems that can directly
receive natural language instructions.

\hypertarget{sec-conclusion}{%
\section{Conclusion}\label{sec-conclusion}}

After decades of unfulfilled aspirations, machine learning research has
recently converged upon a scalable neural network architecture that may
offer a first glimpse of artificial general intelligence. While modern
large language models fall short of human intelligence in significant
ways, they have already become useful tools for domain-general
information processing, and their capabilities have increased at a
staggering pace in the past few years. With increased capabilities comes
an urgent need to ensure these systems are safe and reliable by aligning
them with suitable norms of behaviour.

I have argued that this requirement has not yet been met, and is
unlikely to be met in the near future. While existing alignment
strategies such as fine-tuning and system prompts do reduce the
probability of harmful behaviour during normal use, they can be
circumvented by adversarial attacks that successfully recover such
behaviour. I have offered an explanation for this vulnerability in light
of empirical evidence: adversarial attacks leverage LLMs' remarkable
aptitude to learn (or unlearn) complex behaviours in context,
potentiated by their roleplaying ability and the exploitation of
conflicting norms of alignment.

If this explanation is correct, the susceptibility of LLMs to in-context
misalignment through adversarial attacks is not merely a minor fluke of
current models, but exposes a fundamental trade-off in their design. For
these systems to be useful and versatile, they need to be good at
executing arbitrary user instructions in natural language. This ability
is both a blessing and a curse; it is inherently challenging to robustly
constrain it on alignment-sensitive inputs without thereby undermining
the system's usefulness and versatility.

The implications of this perspective for AI safety should not be
overlooked. Current LLMs can already cause harm. It is likely that
future systems based on similar architectures will have greater
capabilities and therefore greater potential for abuse. If we cannot
solve the value alignment problem for current systems in a way that is
robust to adversarial misalignment, we should be concerned about
developing and deploying future systems that share the same backbone.

My argument also has implications for AI safety regulation. Companies
like OpenAI routinely decline to release the weights of their flagship
models on the basis of stated safety concerns
\citep{altmanPlanningAGI2023}. There is a concurrent push to regulate
the public release of powerful LLMs
\citep{hackerRegulatingChatGPTOther2023}. Given the lack of effective
strategies to fully mitigate in-context misalignment, this regulatory
approach is unlikely to prevent malicious use. Mere access to an API is
sufficient to perform effective adversarial attacks without fine-tuning
the model for nefarious purposes. The trend to increase context length
will continue to blur the line between actual fine-tuning and in-context
learning, and may correspondingly decrease the relevance of publicly
releasing model weights for AI safety. Future regulatory discussions
should take this trend into account, and should not overlook the
significance of the alignment problem in context.

  \bibliography{bibliography.bib}

\end{document}